\documentclass[11pt,twocolumn,letterpaper]{article}

\usepackage{cvpr}
\usepackage{times}
\usepackage{epsfig}
\usepackage{graphicx}
\usepackage{amsmath}
\usepackage{amssymb}
\usepackage[font=small]{caption}
\usepackage[font=small,skip=-2pt]{caption}

\usepackage[breaklinks=true,bookmarks=false]{hyperref}

\cvprfinalcopy 

\def\cvprPaperID{****} 

\begin{document}

\title{Convolutional Neural Networks for Facial Expression Recognition}

\author{Shima Alizadeh\\
Stanford University\\
{\tt\small shima86@stanford.edu}
\and
Azar Fazel\\
Stanford University\\
{\tt\small azarf@stanford.edu}
}

\maketitle

\begin{abstract}
We have developed convolutional neural networks (CNN) for  a facial expression recognition task. The goal is to classify each facial image into one of the seven facial emotion categories considered in this study. We trained CNN models with different depth using gray-scale images. We developed our models in Torch \cite{torch} and exploited Graphics Processing Unit (GPU) computation in order to expedite the training process. In addition to the networks performing based on raw pixel data, we employed a hybrid feature strategy by which we trained a novel CNN model with the combination of raw pixel data and Histogram of Oriented Gradients (HOG) features \cite{hog}. To reduce the overfitting of the models, we utilized different techniques including dropout and batch normalization in addition to L2 regularization. We applied cross validation to determine the optimal hyper-parameters and evaluated the performance of the developed models by looking at their training histories. We also present the visualization of different layers of a network to show what features of a face can be learned by CNN models.  
\end{abstract}

\section{Introduction}
Humans interact with each other mainly through speech, but also through body gestures, to emphasize certain parts of their speech and to display emotions. One of the important ways humans display emotions is through facial expressions which are a very important part of communication. Though nothing is said verbally, there is much to be understood about the messages we send and receive through the use of nonverbal communication. Facial expressions convey non-verbal cues, and they play an important role in interpersonal relations \cite{art,survey}. Automatic recognition of facial expressions can be an important component of natural human-machine interfaces; it may also be used in behavioral science and in clinical practice. Although humans recognize facial expressions virtually without effort or delay, reliable expression recognition by machine is still a challenge. There have been several advances in the past few years in terms of face detection, feature extraction mechanisms and the techniques used for expression classification, but development of an automated system that accomplishes this task is difficult \cite{exp}. In this paper, we present an approach based on Convolutional Neural Networks (CNN) for facial expression recognition. The input into our system is an image; then, we use CNN to predict the facial expression label which should be one these labels: anger, happiness, fear, sadness, disgust and neutral.

\section{Related Work} 
In recent years, researchers have made considerable progress in developing automatic expression classifiers \cite{rw-1,rw-2,rw-3}. Some expression recognition systems classify the face into a set of prototypical emotions such as happiness, sadness and anger.\cite{rw-4}. Others attempt to recognize the individual muscle movements that the face can produce \cite{rw-5} in order to provide an objective description of the face. The best known psychological framework for describing nearly the entirety of facial movements is the Facial Action Coding System (FACS) \cite{facs}. FACS is a system to classify human facial movements by their appearance on the face using Action Units (AU). An AU is one of 46 atomic elements of visible facial movement or its associated deformation; an expression typically results from the accumulation of several AUs \cite{rw-1,rw-2}. Moreover, there have been several developments in the techniques used for facial expression recognition: Bayesian Networks, Neural Networks and the multi-level Hidden Markov Model (HMM) \cite{tech-1,tech-2}. Some of them contain drawbacks of recognition rate or timing. Usually, to achieve accurate recognition two or more techniques can be combined; then, features are extracted as needed. The success of each technique is dependent on pre-processing of the images because of illumination and feature extraction.

\section{Methods} 
We developed CNNs with variable depths to evaluate the performance of these models for facial expression recognition. We considered the following network architecture in our investigation:\\
\begin{center}
[Conv-(SBN)-ReLU-(Dropout)-(Max-pool)]M - [Affine-(BN)-ReLU-(Dropout)]N - Affine - Softmax.
\end{center}

The first part of the network refers to M convolutional layers that can possess spatial batch normalization (SBN), dropout, and max-pooling in addition to the convlution layer and ReLU nonlinearity, which always exists in these layers. After M convolution layers, the network is led to N fully connected layers that always have Affine operation and ReLU nonlinearity, and can include batch normalization (BN) and dropout. Finally, the network is followed by the affine layer that computes the scores and softmax loss function. The developed model gives the user the freedom to decide about the number of convlutional and fully connected layers, as well as the existance of batch normalization, dropout and max-pooling layers. Along with dropout and batch normalization techniques, we included L2 regularization in our implementation. Furthermore, the number of filters, strides, and zero-padding can be specified by user, and if they are not given, the default values are considered.\\
As we will describe in the next section, we proposed the idea of combining HOG features with those extracted by convolutional layers by mean of raw pixel data. To this end, we utilized the same architecture described above, but with this difference that we added the HOG features to those exiting the last convolution layer. The hybrid feature set then enters the fully connected layers for score and loss calculation.\\
We implemented the aforementioned model in Torch and took advantage of GPU accelerated deep learning features to make the model training process faster.

\section{Dataset and Features} 

\begin{figure}[t]
\begin{center}
\includegraphics[width=0.8\linewidth]{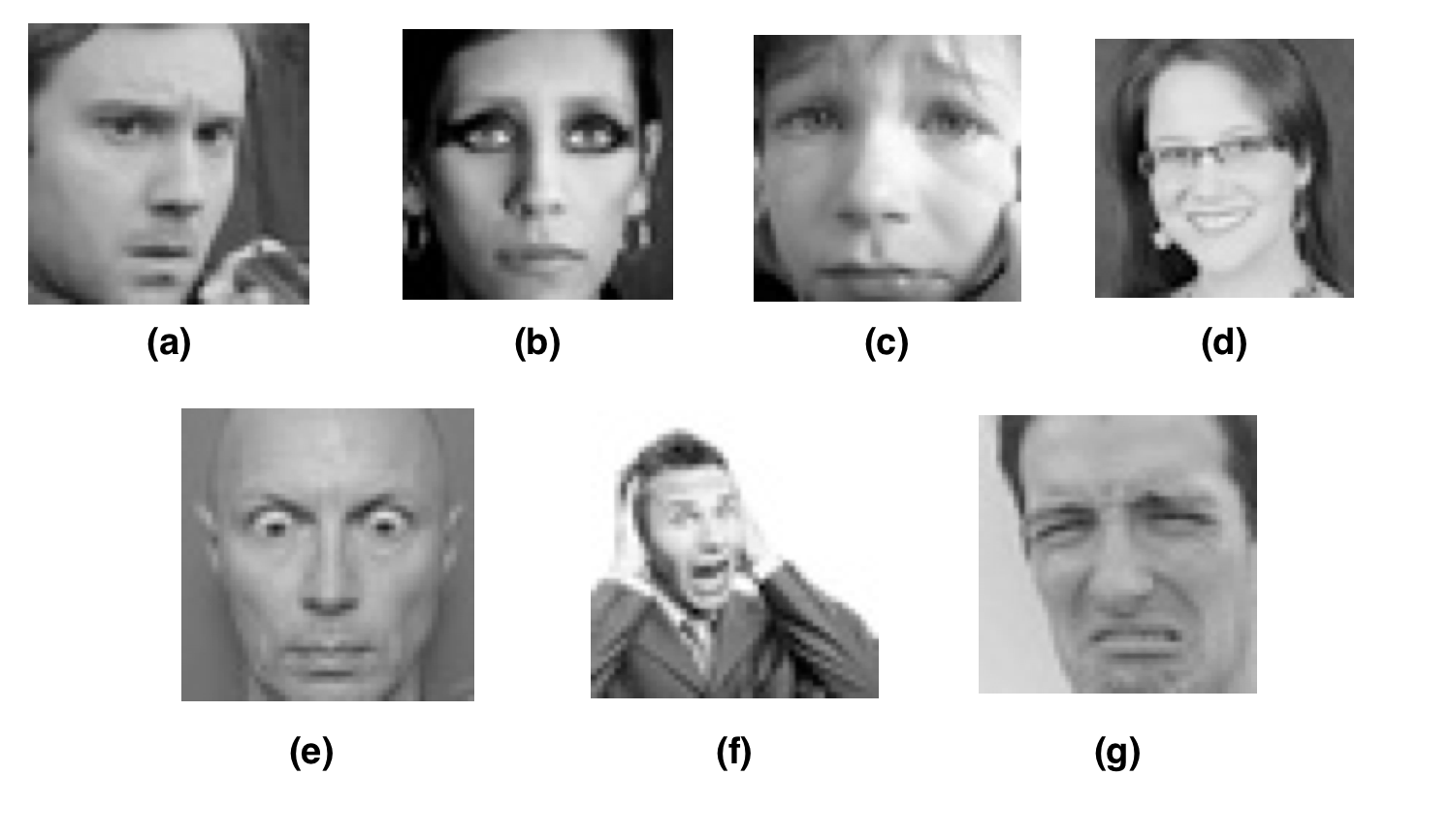}
\end{center}
\caption[font=small]{Examples of seven facial emotions that we consider in this classification problem. (a) angry, (b) neutral, (c) sad, (d) happy, (e) surprise, (f) fear, (g) disgust}
\label{fig:faces}
\end{figure}

In this project, we used a dataset provided by \href{https://www.kaggle.com/c/challenges-in-representation-learning-facial-expression-recognition-challenge/data}{Kaggle website}, which consists of about 37,000 well-structured $48\times 48$ pixel gray-scale images of faces. The images are processed in such a way that the faces are almost centered and each face occupies about the same amount of space in each image. Each image has to be categorized into one of the seven classes that express different facial emotions. These facial emotions have been categorized as: 0=Angry, 1=Disgust, 2=Fear, 3=Happy, 4=Sad, 5=Surprise, and 6=Neutral. Figure \ref{fig:faces} depicts one example for each facial expression category. In addition to the image class number (a number between 0 and 6), the given images are divided into three different sets which are training, validation, and test sets. There are about 29,000 training images, 4,000 validation images, and 4,000 images for testing. After reading the raw pixel data, we normalized them by subtracting the mean of the training images from each image including those in the validation and test sets. For the purpose of data augmentation, we produced mirrored images by flipping images in the training set horizontally.\\
In order to classify the expressions, mainly we used the features generated by convolution layers using the raw pixel data. As an extra exploration, we developed learning models that concatenate the HOG features with those generated by convolutional layers and give them as input features into Fully Connected (FC) layers.

\section{Analysis}
\subsection{Experiments}
For the purpose of this project, first we built a shallow CNN. This network had two convolutional layers and one FC layer. In the first convolutional layer, we had 32 $3\times$3 filters, with the stride of size 1, along with batch normalization and dropout, but without max-pooling. In the second convolutional layer, we had 64 $3\times$3 filters, with the stride od size 1, along with batch normalization and dropout and also max-pooling with a filter size $2\times$2. In the FC layer, we had a hidden layer with 512 neurons and Softmax as the loss function. Also in all the layers, we used Rectified Linear Unit (ReLU) as the activation function. Before training our model, we did some sanity checks to make sure that the implementation of the network was correct. For the first sanity check, we computed the initial loss when there is no regularization. Since our classifier has 7 different classes, we expected to get a value around 1.95. As the second sanity check, we tried to overfit our model using a small subset of the training set. Our shallow model passed both of these sanity checks. Then, we started training our model from scratch. To make the model training process faster, we exploited GPU accelerated deep learning facilities on Torch. For the training process, we used all of the images in the training set with 30 epochs and a batch size of 128 and cross-validated the hyper-patameters of the model with different values for regularization, learning rate and the number of hidden neurons . To validate our model in each iteration, we used the validation set and to evaluate the performance of the model, we used the test set. The best shallow model, gave us 55\% accuracy on the validation set and 54\% on the test set. Table [\ref{table:shallow-params}] summarizes the hyper-parameters obtained by cross validation for the shallow model. 
\begin{table}[t]
\begin{center}
\begin{tabular}{|l|c|}
\hline
Parameter & Value \\
\hline\hline
Learning Rate & 0.001 \\
Regularization & 1e-6 \\
Hidden Neurons & 512\\
\hline
\end{tabular}
\end{center}
\caption{The hyper-parameters obtained by cross validation for the  shallow model}
\label{table:shallow-params}
\end{table}

\begin{figure*}[t]
\begin{center}
\includegraphics[width=14cm]{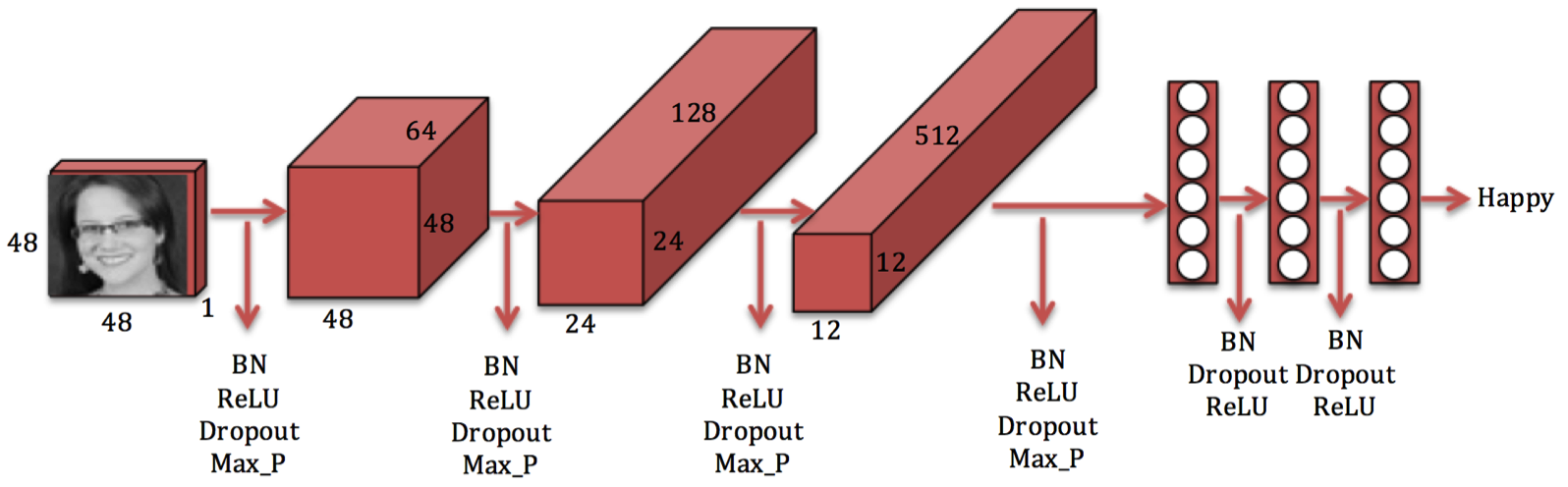}
\end{center}
\caption[font=small]{The architecture of the deep network: 4 convolutional layers and 2 fully connected layers
}
\vspace*{7mm}
\label{fig:deep-model}
\end{figure*}

To observe the effect of adding convolutional layers and FC layers to the network, we trained a deeper CNN with 4 convolutional layers and two FC layers. The first convolutional layer had 64 $3\times$3 filters, the second one had 128 $5\times$5 filters, the third one had 512 $3\times$3 filters and the last one had 512 $3\times$3 filters. In all the convolutional layers, we have  a stride of size 1, batch normalization, dropout, max-pooling and ReLU as the activation function. The hidden layer in the first FC layers had 256 neurons and the second FC layer had 512 neurons. In both FC layers, same as in the convolutional layers, we used batch normalization, dropout and ReLU. Also we used Softmax as our loss function. Figure \ref{fig:deep-model} shows the architecture of this deep network. As in the shallow model, before training the network, we performed initial loss checking and examined the ability of overfitting the network using a small subset of the training set. The results of these sanity checks proved that the implementation of the network was correct. Then, using 35 epochs and a batch size of 128, we trained the network with all the images in the training set. Moreover, we cross-validated the hyper-parameters to get the model with the highest accuracy. This time, we obtained an accuracy of 65\% on the validation set and 64\% on the test set. Table [\ref{table:deep-params}] depicts the values for each hyper-parameter in this model that had the highest accuracy.\\
To explore the deeper CNNs, we also trained networks with 5 and 6 convolutional layers, but they did not increase the classification accuracy. Therefor we considered the model with 4 convolutional layers and 2 FC layers as the best network for our dataset.\\
In both the shallow and deep models, we only exploited the features generated by the convolution layers using the raw pixel data as the main features for our classification task. Usually HOG features are used for facial expression recognition since they are sensitive to edges. We wanted to explore if there is any way to apply HOG features along with raw pixels to our network and observe the performance of the model when it has a combination of two different features. For this, we built a new learning model containing two neural networks: the first one contained convolutional layers, and  the second one had only fully connected layers. The features developed by the first network are concatenated with the HOG features and the resultant hybrid features were fed into the second network. To evaluate the performance of the network with hybrid features, we trained two networks, one shallow network and one deep network with the same characteristics as the shallow and deep networks that we trained for the previous experiment. This time, the accuracy of the shallow model was very close to the accuracy we got from the shallow model that used only raw pixels. The accuracy for the deep model also was similar to the accuracy we got for our deep model with raw pixels, as features.
\begin{table}[t]
\begin{center}
\begin{tabular}{|l|c|}
\hline
Parameter & Value \\
\hline\hline
Learning Rate & 0.01 \\
Regularization & 1e-7 \\
Hidden Neurons & 256, 512\\
\hline
\end{tabular}
\end{center}
\caption{The hyper-parameters obtained by cross validation for the deep model}
\vspace*{7mm}
\label{table:deep-params}
\end{table}
\subsection{Results}
\begin{figure}[t]
\begin{center}
\includegraphics[width=0.8\linewidth]{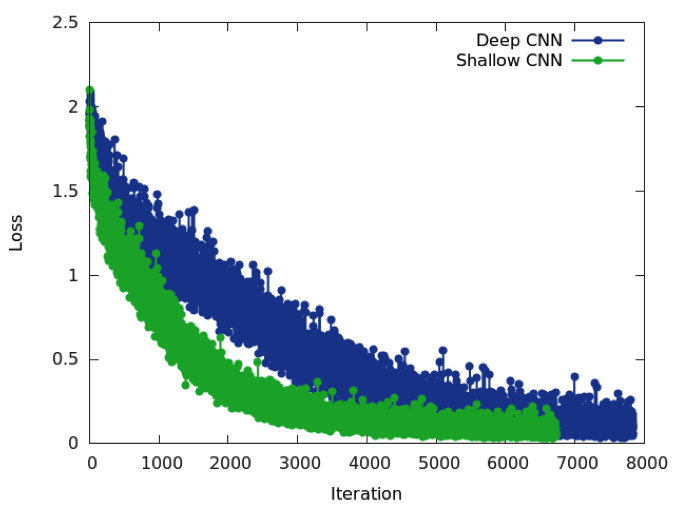}
\end{center}
\caption[font=small]{The loss history of the shallow and deep models}
\vspace*{6mm}
\label{fig:loss}
\end{figure}

\begin{figure}[t]
\begin{center}
\includegraphics[width=0.8\linewidth]{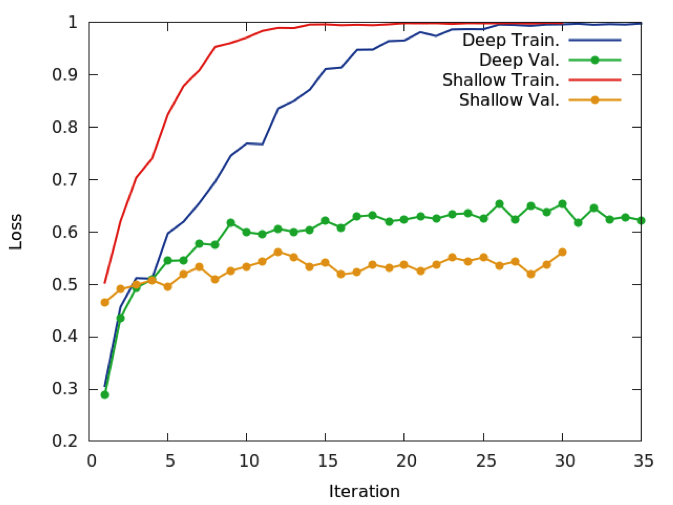}
\end{center}
\caption[font=small]{The accuracy of the shallow and deep models for different numbers of iterations}
\label{fig:acc}
\end{figure}

To compare the performance of the shallow model with the deep model, we plotted the loss history and the obtained accuracy in these models. Figures \ref{fig:loss} and \ref{fig:acc} exhibit the results. As seen in Figure \ref{fig:acc}, the deep network enabled us to increase the validation accuracy by 18.46\%. Furthermore, one can observe that the deep network has reduced the overfitting behavior of the learning model by adding more non-linearity and hierarchical usage of anti-overfitting techniques such as dropout and batch normalization in addition to L2 regularization. From Figure \ref{fig:loss}, we can also see that the shallow network converged faster and the training accuracy quickly reached its highest value.

\begin{figure}[t]
\begin{center}
\includegraphics[width=0.8\linewidth]{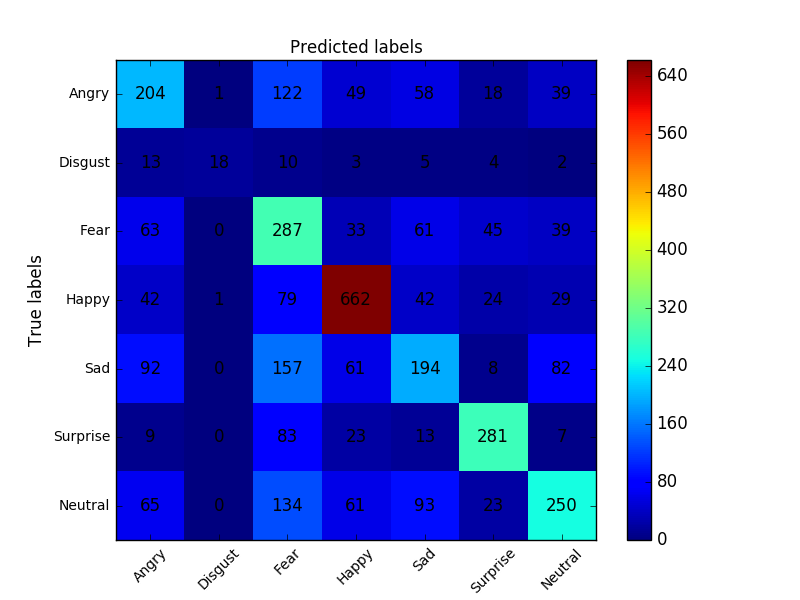}
\end{center}
\caption[font=small]{The confusion matrix for the shallow model}
\label{fig:cm-shallow}
\end{figure}

\begin{figure}[t]
\begin{center}
\includegraphics[width=0.8\linewidth]{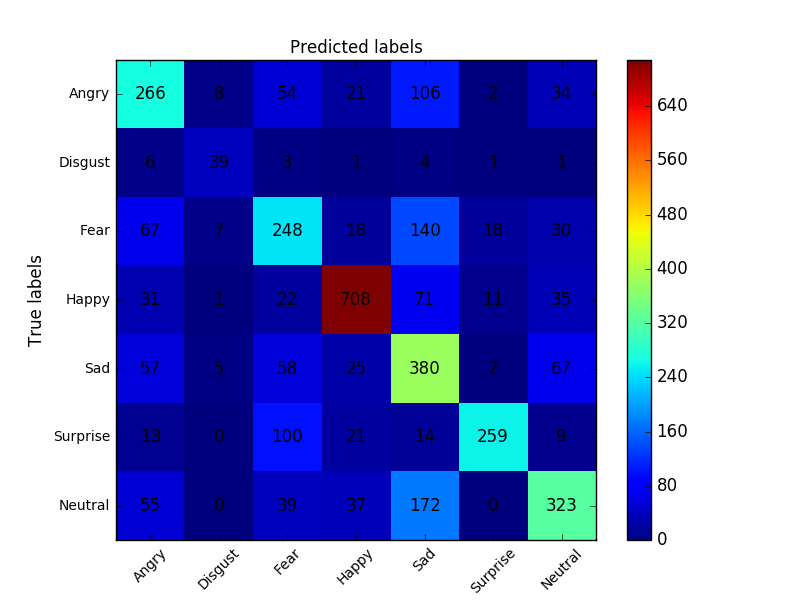}
\end{center}
\caption[font=small]{The confusion matrix for the deep model}
\vspace*{7mm}
\label{fig:cm-deep}
\end{figure}

Moreover, we computed the confusion matrices \cite{scikit} for the shallow and deep networks. Figures \ref{fig:cm-shallow} and \ref{fig:cm-deep} present the visualization of the confusion matrices. As demonstrated in these figures, the deep network results in higher true predictions for most of the labels. It is interesting to see that both models performed well in predicting the happy label, which implies that learning the features of a happy face is easier than other expressions. Additionally, these matrices reveal which labels are likely to be confused by the trained networks. For example, we can see the correlation of angry label with the fear and sad labels. There are lots of instances that their true label is angry but the classifier has misclassified them as fear or sad. These mistakes are consistent with what we see when looking at images in the dataset; even as a human, it can be difficult to recognize whether an angry expression is actually sad or angry. This is due to the fact that people do not all express emotions in the same way.

In addition to confusion matrices, we computed the accuracy of each model for every expression. Table [\ref{table:exp}] shows these results. As seen in this table, the accuracy of predicting happy expression has the highest value among all the emotions in both shallow and deep models. Also, for most of the expressions, using deep network has increased the classification accuracy. For some of the expressions like surprise and fear, using deeper networks not only did not help to get a better accuracy, but also decreased their prediction accuracy. It means that for some expressions, going deeper does not necessarily provide better features.

\begin{figure}[t]
\begin{center}
\includegraphics[width=0.8\linewidth]{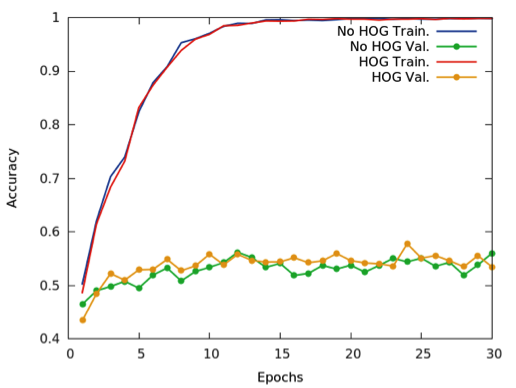}
\end{center}
\caption[font=small]{The accuracy of the shallow model with hybrid features for different numbers of iterations}
\vspace*{7mm}
\label{fig:acc-shallow-hog}
\end{figure}

\begin{figure}[t]
\begin{center}
\includegraphics[width=0.8\linewidth]{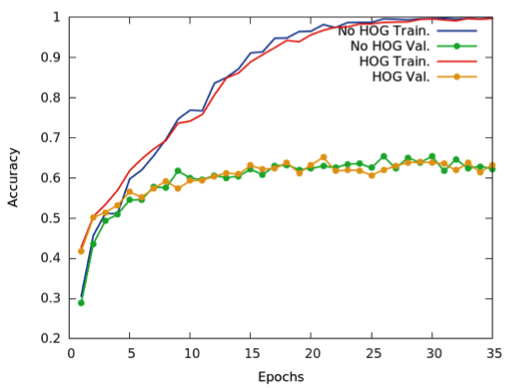}
\end{center}
\caption[font=small]{The accuracy of the deep model with hybrid features for different numbers of iterations}
\vspace*{5mm}
\label{fig:acc-deep-hog}
\end{figure}

\begin{table}[t]
\begin{center}
\begin{tabular}{|c|c|c|}
\hline
Expression & Shallow Model & Deep Model \\
\hline\hline
Angry & 41\% & 53\% \\
Disgust & 32\% & 70\% \\
Fear & 54\% & 46\%\\
Happy & 75\% & 80.5\%\\
Sad & 32\% & 63\%\\
Surprise & 67.5\% & 62.5\%\\
Neutral & 39.9\% & 51.5\%\\
\hline
\end{tabular}
\end{center}
\caption{The accuracy of each expression in the shallow and deep models.}
\vspace*{7mm}
\label{table:exp}
\end{table}
As explained in the previous section, to investigate the effect of utilizing different features in our CNN model, we developed learning models that concatenate the HOG features with those generated by the convolutional layers and used them as input features to the FC layers. Using this idea, we trained one shallow and one deep network. Figure \ref{fig:acc-shallow-hog} and \ref{fig:acc-deep-hog} display the obtained accuracy in different iterations for the shallow and deep models respectively. As seen in these figures, the accuracy of the model is very close to the accuracy we got from the model that has no HOG features. This means that CNN is strong enough to extract sufficient information including those coming from HOG features by using only raw pixel data.

\begin{figure}[t]
\begin{center}
\includegraphics[width=0.8\linewidth]{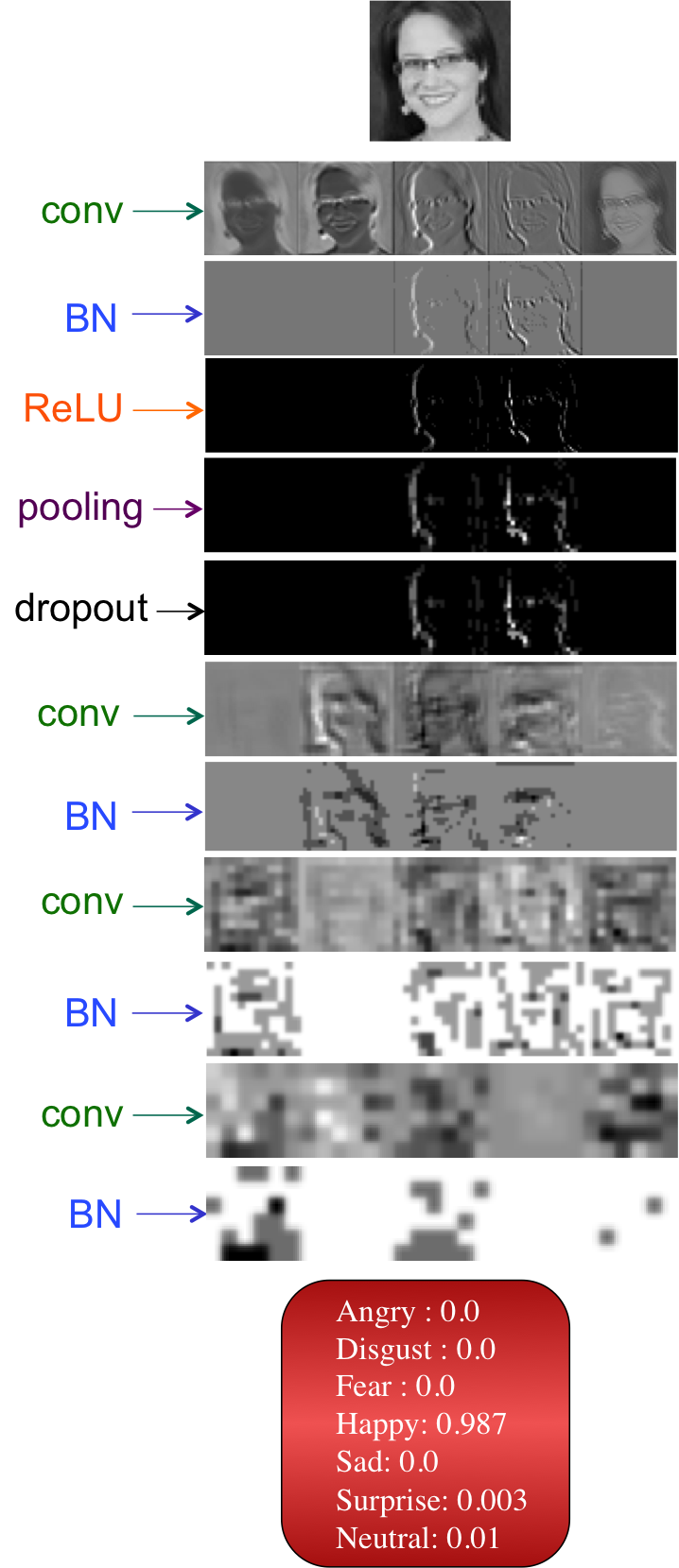}
\end{center}
\caption[font=small]{Visualization of the activation maps for different layers in our CNN}
\label{fig:activation}
\end{figure}

In order to observe the features that our trained network extracts at each layer, we visualized the activation maps of different layers during the forward pass. Figure \ref{fig:activation} shows this visualization. We can see that as the training progresses, the activation maps become more sparse and localized.

\begin{figure}[t]
\begin{center}
\includegraphics[width=0.8\linewidth]{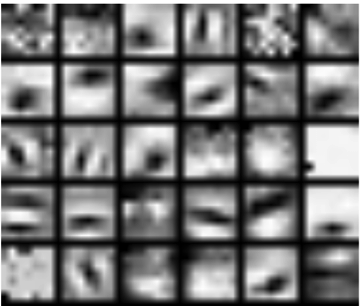}
\end{center}
\caption[font=small]{Visualization of the weights for the first layer in our CNN}
\vspace*{7mm}
\label{fig:weight}
\end{figure}

Furthermore, to see the qualification of the trained network, we also visualized the weights of the first layer. As seen in Figure \ref{fig:weight}, we have smooth filters without any noisy pattern. It indicates that our network has been trained for long enough and the regularization strength is also sufficient.

\begin{figure}[t]
\begin{center}
\includegraphics[width=0.8\linewidth]{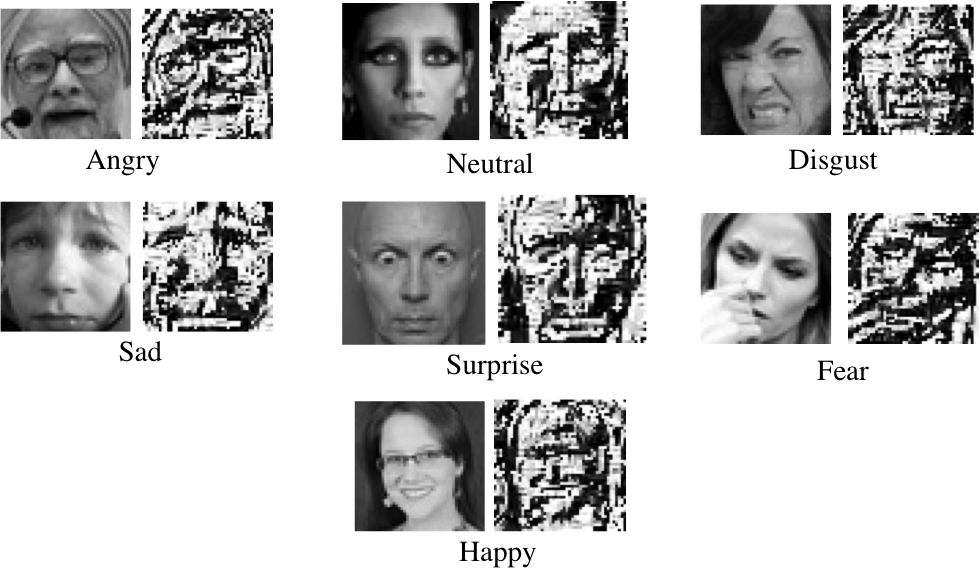}
\end{center}
\caption[font=small]{Examples of applying DeepDream on our dataset}
\label{fig:dreem}
\end{figure}

In addition, We applied DeepDream \cite{deepdream-1,deepdream-2} technique to our best predictive model to find enhanced patterns in our images. Figure \ref{fig:dreem} displays one example for each expression a long with its DeepDream output.

\section{Summary}
\subsection{Conclusion}
We developed various CNNs for a facial expression recognition problem and evaluated their performances using different post-processing and visualization techniques.The results demonstrated that deep CNNs are capable of learning facial characteristics and  improving facial emotion  detection. Also, the hybrid feature sets did not help in improving the model accuracy, which means that the convolutional networks can intrinsically learn the key facial features by using only raw pixel data.
\subsection{Future Work}
For this project, we trained all the models from scratch using CNN packages in Torch. In the future work, we would like to extend our model to color images. This will allow us to investigate the efficacy of pre-trained models such as AlexNet\cite{alex} or VGGNet \cite{vgg} for facial emotion recognition. Another extension would be the implementation of a face detection process, followed by the emotion prediction.


\end{document}